\documentclass{article}

\usepackage[final]{tackling_climate_workshop_style}

\usepackage[utf8]{inputenc} % allow utf-8 input
\usepackage[T1]{fontenc}    % use 8-bit T1 fonts
\usepackage{hyperref}       % hyperlinks
\usepackage{url}            % simple URL typesetting
\usepackage{booktabs}       % professional-quality tables
\usepackage{amsfonts}       % blackboard math symbols
\usepackage{nicefrac}       % compact symbols for 1/2, etc.
\usepackage{microtype}      % microtypography
\usepackage{graphicx}

\title{On the use of Deep Generative Models for ``Perfect'' Prognosis Climate Downscaling}

\author{
  Jose González-Abad\\
  Santander Meteorology Group and \\ Advanced Computing and e-Science Group \\
  Institute of Physics of Cantabria (CSIC-UC) \\
  Santander, Spain \\
  \texttt{gonzabad@ifca.unican.es} \\
  \And
  Jorge Baño-Medina \\
  Santander Meterology Group \\
  Institute of Physics of Cantabria (CSIC-UC) \\
  Santander, Spain \\
  \texttt{bmedina@ifca.unican.es} \\
  \AND
  Ignacio Heredia Cachá \\
  Advanced Computing and e-Science Group \\
  Institute of Physics of Cantabria (CSIC-UC) \\
  Santander, Spain \\
  \texttt{iheredia@ifca.unican.es}
}

\begin{document}

\maketitle

\begin{abstract}
Deep Learning has recently emerged as a ``perfect'' prognosis downscaling technique to compute high-resolution fields from large-scale coarse atmospheric data. Despite their promising results to reproduce the observed local variability, they are based on the estimation of independent distributions at each location, which leads to deficient spatial structures, especially when downscaling precipitation. This study proposes the use of generative models to improve the spatial consistency of the high-resolution fields, very demanded by some sectoral applications (e.g., hydrology) to tackle climate change. 
\end{abstract}

\section{Motivations for generative models in ``perfect'' prognosis downscaling}
Global Climate Models (GCMs) are the main tools used nowadays to study the evolution of climate at different time-scales. They numerically solve a set of equations describing the dynamics of the climate system over a three-dimensional grid (latitude-longitude-height). In climate change modeling, these models are utilized to produce possible future pathways of the climate system based on different natural and anthropogenic forcings. However, due to computational limitations these models present a coarse spatial resolution ---between $1^\circ$ and $3^\circ$,--- which leads to a  misrepresentation of important phenomena occurring at finer scales. The generation of high-resolution climate projections is crucial for important socio-economic activities (e.g., the energy industry), and they are routinely used to elaborate mitigation and adaptation politics to climate change at a regional scale.

Statistical Downscaling (SD) is used to bridge the scale-gap between the coarse model outputs and the local-scale by learning empirical relationships between a set of large-scale variables (predictors) and the regional variable of interest (predictands) based on large simulated/observational historical data records \cite{maraun_statistical_2018}. In this study we focus on a specific type of SD, named the ``Perfect'' Prognosis (PP) approach. PP downscaling leans on observational datasets to learn empirical relationships linking the predictors and the predictands. For the former, reanalysis data ---a global dataset which combines observations with short-range forecasts through data assimilation,--- is typically used, whilst for the latter either high-resolution grids or station-scale records can be employed. Once the relationship is established in these ``perfect'' conditions, we feed the model/algorithm with the equivalent GCM predictor variables to obtain high-resolution climate projections. A wide variety of statistical techniques have been deployed to establish these links, such as (generalized) linear models \cite{gutierrez_intercomparison_2019}, support vector machines \cite{chen_statistical_2010}, random forests \cite{pang_statistical_2017}, classical neural networks \cite{williams_modelling_1998}, and more recently deep learning (DL). In particular, DL has recently emerged as a promising PP technique, showing capabilities to reproduce the observed local climate \cite{pan_improving_2019, bano_configuration_2020, sun_statistical_2021}, whilst showing plausible climate change projections of precipitation and temperature fields over Europe \cite{bano_suitability_2021}. Nonetheless, currently the regression-based nature of most of the existing PP methods, leads to an underestimation of the extremes when the predictors lack from sufficient informative power ---i.e., given a particular predictor configuration there are many possible predictand situations,--- since they output the conditional mean \cite{pryor_differential_2020}. To account for the uncertainty describing the possible extremes is crucial for some activities, and the community has driven its attention to probabilistic regression-based modeling. The probabilistic models used mostly estimate the parameters of selected probability distributions conditioned to the large-scale atmospheric situation. The choice of the distribution depends on the variable of interest to be modeled ---for instance, the temperature follows a Gaussian distribution, whilst wind or precipitation fields present a heavy-tailed structure which better fits with Gamma, Poisson or log-normal density functions,--- and the regression-based models are trained to optimize the negative log-likelihood of the selected distribution at each site \cite{williams_modelling_1998, bano_configuration_2020, carreau_stochastic_2011, vaughan_multivariate_2021, vaughan_convolutional_2021}. To model the spatial dependencies among sites, ideally we would estimate multivariate distributions representing the whole predictand domain, instead of predicting independent probability functions at each predictand site. Nonetheless, this was in practice computationally intractable, and very few procedures aimed to downscale over low-dimensional predictand spaces have been successfully deployed \cite{cannon_probabilistic_2008, ben_probabilistic_2014, alaya_non_2018}. 

Recently, deep generative models have been developed that seek to approximate high-dimensional distributions through DL topologies. Based on previous merits in other disciplines, such as image-super-resolution (see e.g., \cite{dong_image_2015, li_feedback_2019}), some studies have searched for an analogy between this task and downscaling, deploying Generative Adversarial Networks (GAN, \cite{leinonen_stochastic_2020, franccois_adjusting_2021}) to obtain stochastic samples of high-resolution precipitation and temperature fields conditioned to their counterpart low-resolution ones. Despite these first studies are far from the PP approach, ---since they lean on surface variables in their predictor set, which are not well represented by GCMs (see \cite{maraun_statistical_2018, maraun_statistical_2019} for guidelines/details on PP),--- they show the potential of generative models to attain impressive levels of spatial structure in their stochastic downscaled predictions. Following this idea, we state that these topologies may provide a tractable alternative to model multivariate conditional distributions over high-dimensional domains in a PP setting, providing stochastic and spatially consistent downscaled fields very demanded by some sectoral applications for climate impact studies. To prove the potential of this type of DL topologies for PP-based downscaling, we show in the next section a use-case where Conditional Variational Auto-Encoders (CVAE) are deployed to produce stochastic high-resolution precipitation fields over Europe.

\section{A downscaling case study over Europe with CVAE} \label{s:CVAE}

We develop a simple use-case \footnote{The code of the use-case is available at https://github.com/jgonzalezab/CVAE-PP-Downscaling} which seeks to illustrate the promising capabilities of CVAE topologies to generate spatially consistent stochastic downscaled fields, especially as compared to the recent state-of-the-art PP DL-based topologies, which are based on the estimation of conditional Bernoulli-Gamma distributions at each predictand site (we refer the reader to \cite{bano_configuration_2020} for more details). To this aim, we deploy the CVAE in the same conditions than \cite{bano_configuration_2020}, which builds on the validation framework proposed in the COST action VALUE \cite{maraun_value_2015}. VALUE proposes the use of ERA-Interim \cite{dee_era_2011} reanalysis variables as predictors ---trimmed to an horizontal resolution of $2^o$,---  and the regular gridded $0.5^o$ E-OBS dataset \cite{cornes_ensemble_2018} as predictand. For the predictor set we use ﬁve thermodynamical variables (geopotential height, zonal and meridional wind, temperature, and specific humidity) at four different vertical levels (1000, 850, 700 and 500 hPa), whilst as predictand we use the daily accumulated precipitation over Europe. The models are trained on the period 1979-2002 and tested on 2003-2008. 

\begin{figure}[h!]
  \centering
  \includegraphics[width = \textwidth]{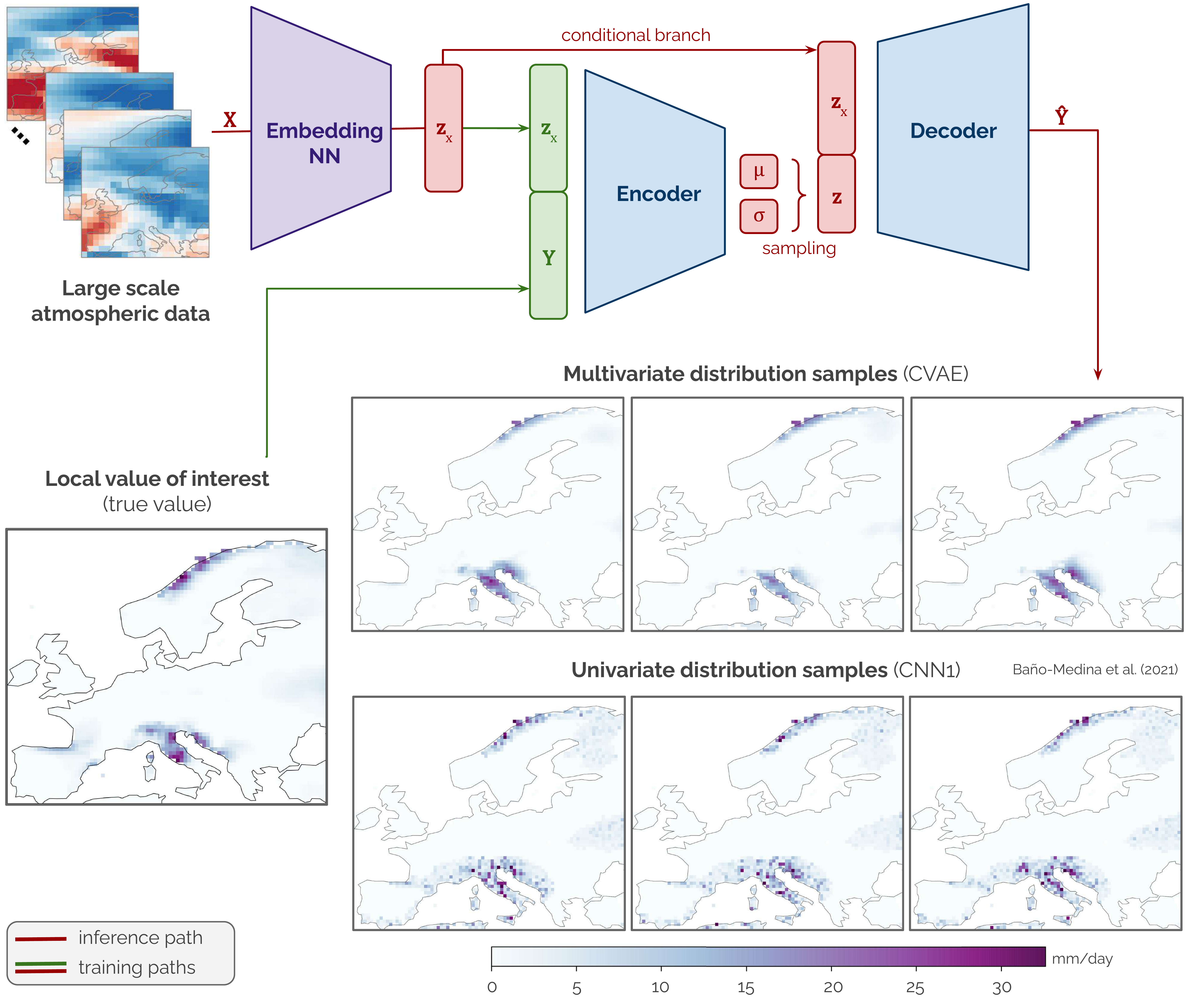}
  \caption{CVAE model architecture. Red lines represent the path followed by the model in the inference phase, during training both paths (green and red lines) are covered. At the bottom, a comparison between three different downscaled fields sampled from CVAE and CNN1 models, alongside the actual observation for 19/02/2004.}
  \label{CVAE}
\end{figure}

Figure \ref{CVAE} shows the scheme of the CVAE proposed. This models builds on three different neural networks ---an embedding network, an encoder and a decoder,--- to produce stochastic samples of precipitation by sampling from a latent distribution which represents the complex interactions between predictors and predictands. During training, the embedding network transforms the high-dimensional predictors $X$ to a low-dimensional array $\mathbf{z}_x$. This array is then stacked with the high-resolution predictand fields $Y$ to feed the encoder network. The encoder outputs the parameters of a Gaussian distribution (i.e., the mean $\mu$ and the standard deviation $\sigma$), which encodes the spatial dependencies between both predictor and predictand fields. During both training and inference phases, stochastic realizations $\mathbf{z}$ sampled from this latent distribution are stacked with the low-dimensional predictor's embedding $\mathbf{z}_x$. This is used to feed the decoder network, which outputs the precipitation values $\widehat{Y}$ at each E-OBS predictand site considered. Therefore, different samples $\widehat{Y}$ conditioned on the same large-scale atmospheric situation $X$ can be generated by sampling different vectors $\mathbf{z}$ from the latent distribution (see the three maps obtained for a particular day). We refer the reader to \cite{sohn_learning_2015} for more details on CVAE.

For the sake of comparison, we select CNN1, which was one of the models that ranked first in \cite{bano_configuration_2020}, as an example of univariate model and compare its stochastic downscaled fields with those of CVAE. It can be seen how CNN1 fields present a spotty structure, characteristic of the sampling performed over the independent Bernoulli-Gamma distributions at each E-OBS site. In contrast, CVAE does not suffer from this problem improving the spatial consistency of the downscaled fields, as can be seen in the smoothness of the predictions.

\section{Pathway of generative models to tackle climate change}
Overall, we have showed the ability of CVAEs to produce spatially consistent stochastic fields in PP setups on a use-case over Europe. The generation of these high-resolution fields through generative models may foster the use of this type of downscaling into climate impact studies, since their products are very demanded by different sectors (e.g., agriculture, hydrology) to tackle climate change. In this line there are several challenges to address. For instance further research is needed in the evaluation of these models on aspects such as temporal consistency, and reproducibility of extremes. Also, in order to apply them to climate change projections, a study of its extrapolation capabilities is also required. The CVAE model developed here is a first approach, but further tuning this architecture may translate in improvements in the generated downscaled fields. For example, \cite{rezende_variational_2015, kingma_improved_2016} propose the use of normalizing flows to generate more complex latent distributions which could help capturing the complex non-linearities of the distribution of precipitation fields. Finally, the DL ecosystem offers a wide catalog of additional topologies which are of interest for PP downscaling (e.g., Conditional GANs \cite{mirza_conditional_2014}).

\par{\bf Acknowledgements.}
The authors acknowledge support from Universidad de Cantabria and Consejería de Universidades, Igualdad, Cultura y Deporte del Gobierno de Cantabria via the ``instrumentación y ciencia de datos para sondear la naturaleza del universo'' project. J. González-Abad would also like to acknowledge the support of the funding from the Spanish \textit{Agencia Estatal de Investigación}
through the \textit{Unidad de Excelencia María de Maeztu} with reference MDM-2017-0765.

\bibliographystyle{unsrt} 
\bibliography{bibliography}

\begin{thebibliography}{10}

\bibitem{maraun_statistical_2018}
Douglas Maraun and Martin Widmann.
\newblock {\em Statistical downscaling and bias correction for climate
  research}.
\newblock Cambridge University Press, 2018.

\bibitem{gutierrez_intercomparison_2019}
Jos{\'e}~Manuel Guti{\'e}rrez, Douglas Maraun, Martin Widmann, Radan Huth, Elke
  Hertig, Rasmus Benestad, Ole R{\"o}ssler, Joanna Wibig, Renate Wilcke, Sven
  Kotlarski, et~al.
\newblock An intercomparison of a large ensemble of statistical downscaling
  methods over europe: Results from the value perfect predictor
  cross-validation experiment.
\newblock {\em International journal of climatology}, 39(9):3750--3785, 2019.

\bibitem{chen_statistical_2010}
Shien-Tsung Chen, Pao-Shan Yu, and Yi-Hsuan Tang.
\newblock Statistical downscaling of daily precipitation using support vector
  machines and multivariate analysis.
\newblock {\em Journal of hydrology}, 385(1-4):13--22, 2010.

\bibitem{pang_statistical_2017}
Bo~Pang, Jiajia Yue, Gang Zhao, and Zongxue Xu.
\newblock Statistical downscaling of temperature with the random forest model.
\newblock {\em Advances in Meteorology}, 2017, 2017.

\bibitem{williams_modelling_1998}
Peter~M Williams.
\newblock Modelling seasonality and trends in daily rainfall data.
\newblock In {\em Advances in neural information processing systems}, pages
  985--991, 1998.

\bibitem{pan_improving_2019}
Baoxiang Pan, Kuolin Hsu, Amir AghaKouchak, and Soroosh Sorooshian.
\newblock Improving precipitation estimation using convolutional neural
  network.
\newblock {\em Water Resources Research}, 55(3):2301--2321, 2019.

\bibitem{bano_configuration_2020}
Jorge Ba{\~n}o-Medina, Rodrigo Manzanas, and Jos{\'e}~Manuel Guti{\'e}rrez.
\newblock Configuration and intercomparison of deep learning neural models for
  statistical downscaling.
\newblock {\em Geoscientific Model Development}, 13(4):2109--2124, 2020.

\bibitem{sun_statistical_2021}
Lei Sun and Yufeng Lan.
\newblock Statistical downscaling of daily temperature and precipitation over
  china using deep learning neural models: Localization and comparison with
  other methods.
\newblock {\em International Journal of Climatology}, 41(2):1128--1147, 2021.

\bibitem{bano_suitability_2021}
Jorge Ba{\~n}o-Medina, Rodrigo Manzanas, and Jos{\'e}~Manuel Guti{\'e}rrez.
\newblock On the suitability of deep convolutional neural networks for
  continental-wide downscaling of climate change projections.
\newblock {\em Climate Dynamics}, pages 1--11, 2021.

\bibitem{pryor_differential_2020}
SC~Pryor and JT~Schoof.
\newblock Differential credibility assessment for statistical downscaling.
\newblock {\em Journal of Applied Meteorology and Climatology},
  59(8):1333--1349, 2020.

\bibitem{carreau_stochastic_2011}
Julie Carreau and Mathieu Vrac.
\newblock Stochastic downscaling of precipitation with neural network
  conditional mixture models.
\newblock {\em Water Resources Research}, 47(10), 2011.

\bibitem{vaughan_multivariate_2021}
Anna Vaughan, Nicholas~D Lane, and Michael Herzog.
\newblock Multivariate climate downscaling with latent neural processes.
\newblock 2021.

\bibitem{vaughan_convolutional_2021}
Anna Vaughan, Will Tebbutt, J~Scott Hosking, and Richard~E Turner.
\newblock Convolutional conditional neural processes for local climate
  downscaling.
\newblock {\em Geoscientific Model Development Discussions}, pages 1--25, 2021.

\bibitem{cannon_probabilistic_2008}
Alex~J Cannon.
\newblock Probabilistic multisite precipitation downscaling by an expanded
  bernoulli--gamma density network.
\newblock {\em Journal of Hydrometeorology}, 9(6):1284--1300, 2008.

\bibitem{ben_probabilistic_2014}
Mohamed~Ali Ben~Alaya, Fateh Chebana, and Taha~BMJ Ouarda.
\newblock Probabilistic gaussian copula regression model for multisite and
  multivariable downscaling.
\newblock {\em Journal of Climate}, 27(9):3331--3347, 2014.

\bibitem{alaya_non_2018}
MA~Ben Alaya, Taha~BMJ Ouarda, and Fateh Chebana.
\newblock Non-gaussian spatiotemporal simulation of multisite daily
  precipitation: downscaling framework.
\newblock {\em Climate dynamics}, 50(1):1--15, 2018.

\bibitem{dong_image_2015}
Chao Dong, Chen~Change Loy, Kaiming He, and Xiaoou Tang.
\newblock Image super-resolution using deep convolutional networks.
\newblock {\em IEEE transactions on pattern analysis and machine intelligence},
  38(2):295--307, 2015.

\bibitem{li_feedback_2019}
Zhen Li, Jinglei Yang, Zheng Liu, Xiaomin Yang, Gwanggil Jeon, and Wei Wu.
\newblock Feedback network for image super-resolution.
\newblock In {\em Proceedings of the IEEE/CVF Conference on Computer Vision and
  Pattern Recognition}, pages 3867--3876, 2019.

\bibitem{leinonen_stochastic_2020}
Jussi Leinonen, Daniele Nerini, and Alexis Berne.
\newblock Stochastic super-resolution for downscaling time-evolving atmospheric
  fields with a generative adversarial network.
\newblock {\em IEEE Transactions on Geoscience and Remote Sensing}, 2020.

\bibitem{franccois_adjusting_2021}
Bastien Fran{\c{c}}ois, Soulivanh Thao, and Mathieu Vrac.
\newblock Adjusting spatial dependence of climate model outputs with
  cycle-consistent adversarial networks.
\newblock {\em Climate Dynamics}, pages 1--31, 2021.

\bibitem{maraun_statistical_2019}
Douglas Maraun, Martin Widmann, and Jos{\'e}~M Guti{\'e}rrez.
\newblock Statistical downscaling skill under present climate conditions: A
  synthesis of the value perfect predictor experiment.
\newblock {\em International Journal of Climatology}, 39(9):3692--3703, 2019.

\bibitem{maraun_value_2015}
Douglas Maraun, Martin Widmann, Jos{\'e}~M Guti{\'e}rrez, Sven Kotlarski,
  Richard~E Chandler, Elke Hertig, Joanna Wibig, Radan Huth, and Renate~AI
  Wilcke.
\newblock Value: A framework to validate downscaling approaches for climate
  change studies.
\newblock {\em Earth's Future}, 3(1):1--14, 2015.

\bibitem{dee_era_2011}
Dick~P Dee, S~M Uppala, AJ~Simmons, Paul Berrisford, P~Poli, S~Kobayashi,
  U~Andrae, MA~Balmaseda, G~Balsamo, d~P Bauer, et~al.
\newblock The era-interim reanalysis: Configuration and performance of the data
  assimilation system.
\newblock {\em Quarterly Journal of the royal meteorological society},
  137(656):553--597, 2011.

\bibitem{cornes_ensemble_2018}
Richard~C. Cornes, Gerard van~der Schrier, Else J. M. van~den Besselaar, and
  Philip~D. Jones.
\newblock An {Ensemble} {Version} of the {E}-{OBS} {Temperature} and
  {Precipitation} {Data} {Sets}.
\newblock {\em Journal of Geophysical Research: Atmospheres},
  123(17):9391--9409, 2018.

\bibitem{sohn_learning_2015}
Kihyuk Sohn, Honglak Lee, and Xinchen Yan.
\newblock Learning structured output representation using deep conditional
  generative models.
\newblock {\em Advances in neural information processing systems},
  28:3483--3491, 2015.

\bibitem{rezende_variational_2015}
Danilo Rezende and Shakir Mohamed.
\newblock Variational inference with normalizing flows.
\newblock In {\em International conference on machine learning}, pages
  1530--1538. PMLR, 2015.

\bibitem{kingma_improved_2016}
Durk~P Kingma, Tim Salimans, Rafal Jozefowicz, Xi~Chen, Ilya Sutskever, and Max
  Welling.
\newblock Improved variational inference with inverse autoregressive flow.
\newblock {\em Advances in neural information processing systems},
  29:4743--4751, 2016.

\bibitem{mirza_conditional_2014}
Mehdi Mirza and Simon Osindero.
\newblock Conditional generative adversarial nets.
\newblock {\em arXiv preprint arXiv:1411.1784}, 2014.

\end{thebibliography}

\end{document}